\documentclass{article}

\usepackage[preprint]{corl_2026} 
\usepackage{amsmath}
\usepackage{amssymb}
\usepackage{booktabs}
\usepackage{multirow}
\usepackage[table]{xcolor}
\usepackage{makecell}
\usepackage{graphicx}
\usepackage[american]{babel}
\usepackage{microtype}
\usepackage{enumitem}
\usepackage{wrapfig}
\usepackage{needspace}

\makeatletter
\renewcommand{\section}{\@startsection{section}{1}{\z@}%
  {-1.2ex \@plus -0.5ex \@minus -0.2ex}%
  { 0.6ex \@plus  0.2ex \@minus  0.1ex}%
  {\large\bf\raggedright}}
\renewcommand{\subsection}{\@startsection{subsection}{2}{\z@}%
  {-1.0ex \@plus -0.5ex \@minus -0.2ex}%
  { 0.3ex \@plus  0.1ex}%
  {\normalsize\bf\raggedright}}
\renewcommand{\subsubsection}{\@startsection{subsubsection}{3}{\z@}%
  {-0.8ex \@plus -0.4ex \@minus -0.2ex}%
  { 0.2ex \@plus  0.1ex}%
  {\normalsize\bf\raggedright}}
\renewcommand{\paragraph}{\@startsection{paragraph}{4}{\z@}%
  {0.8ex \@plus 0.3ex \@minus 0.2ex}%
  {-0.6em}%
  {\normalsize\bf}}
\makeatother

\setlist{topsep=2pt,partopsep=0pt,itemsep=1pt,parsep=1pt,leftmargin=*}

\usepackage[skip=4pt]{caption}
\title{Where Should Action Generation Begin?\\A Learnable Source Prior for Generative Robot Policies}

\author{
  \textbf{Meipo Dai$^{*,1}$ \quad Qiyuan Zhuang$^{*,1}$ \quad He-Yang Xu$^{*,1}$ \quad Ying-Jie Shuai$^{1}$} \\
  \textbf{Yijun Wang$^{1}$ \quad Qi Dou$^{2}$ \quad Xiu-Shen Wei$^{\dagger,1}$} \\
  $^{1}$Southeast University, \quad $^{2}$The Chinese University of Hong Kong \\
  {\small $^{*}$Equal contribution. \quad $^{\dagger}$Corresponding author.}
}

\begin{document}
\maketitle


\begin{abstract}
Generative robot policies typically begin action generation from an observation-independent standard Gaussian distribution, leaving the choice of source distribution underexplored. This work asks a simple question: \emph{where should action generation begin}? We propose LeaP, a \underline{Lea}rnable source \underline{P}rior that replaces the standard Gaussian with a proprioception-conditioned diagonal Gaussian over action chunks. Parameterized by a lightweight MLP, LeaP jointly predicts the mean and state-adaptive variance of the source distribution, while keeping the downstream generator architecture and inference solver unchanged. This design provides an observation-informed yet stochastic initialization, allowing the generator to focus on precise action refinement rather than transporting samples from an uninformed noise source. On 15 RoboTwin manipulation tasks, LeaP achieves an average success rate of 81.6\%, outperforming four representative baselines---including deterministic-source methods, a no-prior counterpart, and a diffusion-bridge policy---by 6.5 to 25.5 percentage points. The same prior consistently improves both flow-matching and diffusion-bridge generators, while using fewer parameters and converging faster. The advantage carries over to real-world deployment, where LeaP attains the best performance. These results suggest that the source distribution is an independent and reusable design axis for generative robot policies, complementary to the choice of generative dynamics.
\end{abstract}


\section{Introduction}

\begin{figure}[t]
\centering
\includegraphics[width=\linewidth]{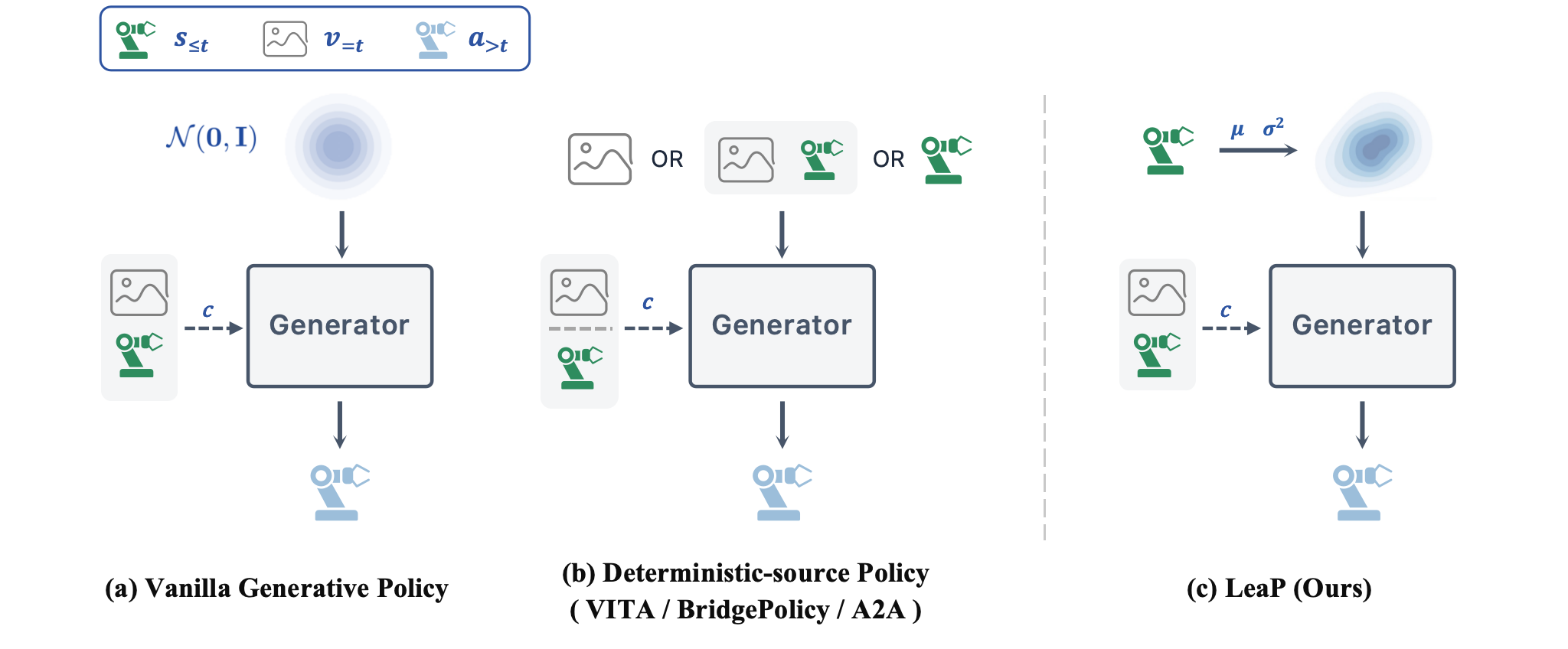}
\caption{\textbf{Three strategies for the source distribution in generative robot policies.}
\textbf{(a)} Vanilla policies sample from an observation-independent $\mathcal{N}(\mathbf{0},\mathbf{I})$, leaving the generator to transport from an uninformed source to observation-relevant actions.
\textbf{(b)} Deterministic-source policies (VITA, BridgePolicy, A2A) replace the standard Gaussian with a deterministic, observation-derived source, but leave its uncertainty unmodeled.
\textbf{(c)} LeaP models the source as a proprioception-conditioned, \emph{learnable} diagonal Gaussian, jointly predicting the mean $\boldsymbol{\mu}$ and a state-adaptive variance $\boldsymbol{\sigma}^2$, providing an observation-informed yet stochastic initialization.}
\vspace{-0.8em}
\label{fig:teaser}
\end{figure}

Visuomotor policies have emerged as a powerful paradigm for robot imitation learning~\citep{chi2023diffusionpolicy, dp3, flowpolicy, ho2020ddpm, lipman2023flow, pi0}. 
By casting action generation as conditional sampling, diffusion-based~\citep{ho2020ddpm, chi2023diffusionpolicy, dp3} and flow-matching policies~\citep{lipman2023flow, flowpolicy, pi0} can model rich action distributions from expert demonstrations. 
Despite their architectural diversity, as depicted in Fig.~\ref{fig:teaser}(a), these methods typically share the same default starting point: action generation begins from an observation-independent standard Gaussian $\mathcal{N}(\mathbf{0}, \mathbf{I})$~\citep{chi2023diffusionpolicy, dp3, flowpolicy, lipman2023flow}. 
Such a source distribution contains no information about the current robot state or task context. 
Consequently, the generator must spend part of its integration budget transporting samples from an uninformed source to task-relevant actions.
This raises a simple but underexplored question: \emph{where should action generation begin?}

Recent studies have started to challenge the standard Gaussian source by deriving the starting point from observations. As illustrated in Fig.~\ref{fig:teaser}(b), this line of work replaces the standard Gaussian with a deterministic, observation-derived source: A2A~\citep{jia2026action} constructs the source from historical proprioceptive sequences, VITA~\citep{gao2025vita} uses visual latents, and BridgePolicy~\citep{bridgepolicy} incorporates visual and proprioceptive observations into the source trajectory of a diffusion bridge. 
These methods differ in the observation modality and generative formulation, but they largely treat the source as an observation-derived point estimate or a tightly specified observation-conditioned initialization. 
As a result, the uncertainty of the source remains overlooked in generative robot policies. 
This issue is particularly relevant in light of recent analysis by \citet{pan2025much}, which suggests that the strength of generative robot policies stems not merely from distributional expressivity, but from iterative computation and stochasticity injection. 
Although A2A explores injecting fixed-variance Gaussian noise into its proprioceptive input, this heuristic improves robustness at the cost of nominal-task performance and does not explicitly model source uncertainty. 
Thus, the source distribution remains an underexplored design axis for generative robot policies: \emph{should the source be deterministic or distributional? If it is distributional, should its variance be fixed or state-adaptive? Which observation modality should parameterize it?}

We address these questions with LeaP (\underline{Lea}rnable source \underline{P}rior), a proprioception-conditioned learnable source distribution that replaces the standard Gaussian $\mathcal{N}(\mathbf{0}, \mathbf{I})$ as the starting point of action generation. 
As depicted in Fig.~\ref{fig:teaser}(c), LeaP uses a lightweight MLP to map the proprioceptive state feature to the mean and state-adaptive variance of a diagonal Gaussian over the action space.
The prior is trained end-to-end together with the generator using flow matching, negative log-likelihood (NLL), and a CLIP-style symmetric contrastive alignment loss~\citep{radford2021clip, oord2018cpc}. 
At inference time, the source sample is drawn from this proprioception-conditioned Gaussian rather than from observation-agnostic noise, providing an initialization that is both observation-informed and stochastic. 
This creates a natural division of labor: the prior places the source sample in a proprioceptively plausible region of the action space, while the generator performs precise observation-conditioned refinement toward the expert action. 
Importantly, LeaP is plug-and-play: it modifies only the source distribution, leaving the generator architecture, generative dynamics, and inference solver unchanged.

We conduct a systematic empirical study on 15 manipulation tasks from RoboTwin 2.0~\citep{chen2025robotwin}. 
Our key findings are:
\textbf{(i)} LeaP achieves an average success rate of 81.6\%, outperforming four representative baselines---deterministic-source methods A2A~\citep{jia2026action} and VITA~\citep{gao2025vita}, a no-prior counterpart, and the diffusion-bridge policy BridgePolicy~\citep{bridgepolicy}---by 6.5 to 25.5 percentage points.
\textbf{(ii)} Beyond accuracy, LeaP uses a smaller parameter budget and exhibits faster and more stable training convergence.
\textbf{(iii)} Among input modalities, proprioceptive state alone is sufficient to parameterize the prior, while incorporating visual features degrades performance in our setting.
\textbf{(iv)} Among prior parameterizations, jointly learning the mean and a state-adaptive variance is the key factor, outperforming deterministic, variance-agnostic, full-covariance, and mixture-based alternatives.
Together, these results suggest that the source distribution is an independent and reusable design axis for generative robot policies, complementary to the choice of generative dynamics.

\section{Related Work}
\label{sec:related}

\paragraph{Visuomotor policies} Visuomotor policies map multi-modal observations to low-level control actions. While early methods such as ACT~\citep{zhao2023act} learn deterministic observation-to-action mappings, recent work has increasingly adopted generative formulations~\citep{chi2023diffusionpolicy, dp3, flowpolicy, prasad2024consistency, pi0}. Diffusion Policy~\citep{chi2023diffusionpolicy} formulates action generation as iterative denoising, building on diffusion and score-based generative models~\citep{ho2020ddpm, song2021scoresde, karras2022edm}. Flow matching~\citep{lipman2023flow, liu2023rectified, albergo2023interpolant} further provides a simpler generative framework and has been adopted for robot policy learning~\citep{flowpolicy, hu2024adaflow, mp1, yan2025maniflow, zhang2024affordance, chisari2024learning} and large-scale vision-language-action models~\citep{pi0, intelligence2025pi05, gr00t, smolvla}. Despite their architectural diversity, these methods typically initialize action generation from an observation-independent standard Gaussian $\mathcal{N}(\mathbf{0}, \mathbf{I})$, leaving the entire transport from this uninformed source to the action distribution to the generator.

\paragraph{Source distribution design}
Recent image-generation work replaces $\mathcal{N}(\mathbf{0},\mathbf{I})$ with a learned or informed initialization~\citep{mao2024lottery, samuel2024generating, eyring2024reno, ahn2024noiseworthdiffusionguidance}, though for open-loop synthesis rather than closed-loop robot policy learning. In robotics, the source has received comparatively little attention. DSRL~\citep{dsrl} runs RL over the latent-noise space of a frozen policy but learns no standalone source. A more direct line replaces the standard Gaussian with an observation-derived source: a deterministic point estimate from proprioceptive sequences (A2A~\citep{jia2026action}) or a visual latent (VITA~\citep{gao2025vita}); a narrow Gaussian centered on the previously executed action whose variance is a fixed hyperparameter zeroed at test time (Streaming Flow Policy~\citep{jiang2025streaming}); or a vision--state fusion coupled with a diffusion-bridge generator (BridgePolicy~\citep{bridgepolicy}). Across these methods the source is a deterministic point estimate or a fixed, hand-specified distribution, sometimes coupled with a generator redesign. We instead model the source as a proprioception-conditioned, learnable diagonal Gaussian, jointly learning the mean and a state-adaptive variance; this lightweight, plug-in prior stays stochastic at inference and leaves the downstream generator entirely unchanged.

\section{Methodology}
\label{sec:method}
\begin{figure*}[t]
    \centering
    \includegraphics[width=0.9\linewidth]{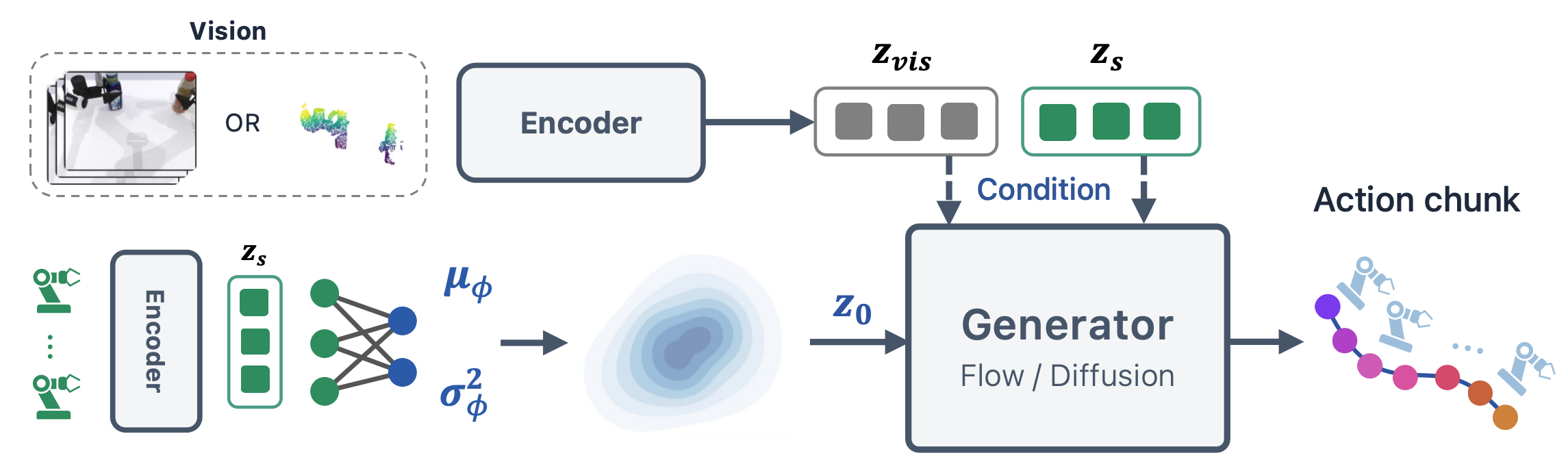}
    \caption{\textbf{Overview of LeaP.}
    The observation is encoded into a visual feature $\mathbf{z}_\text{vis}$ and a proprioceptive feature $\mathbf{z}_s$.
    The prior network $g_\phi$ takes $\mathbf{z}_s$ alone and predicts a proprioception-conditioned diagonal Gaussian, from which the source sample $\mathbf{z}_0$ is drawn via reparameterization.
    The flow-matching generator $v_\theta$ then transports $\mathbf{z}_0$ to the target action under the full conditioning $\mathbf{z}_\text{cond} = [\mathbf{z}_\text{vis};\,\mathbf{z}_s]$.
    Both modules are trained jointly via $\mathcal{L}_\text{flow} + \beta\,\mathcal{L}_\text{NLL} + \alpha\,\mathcal{L}_\text{align}$.}
    \vspace{-0.8em}
    \label{fig:overview}
\end{figure*}
\subsection{Recap of Flow Matching for Policy Learning}
We briefly recap flow matching~\cite{lipman2023flow} as the generative formulation
adopted in this work. Consider visuomotor imitation learning from a dataset of
expert demonstrations $\mathcal{D} = \{(\mathbf{o}_i, \mathbf{a}_i)\}_{i=1}^{N}$,
where $\mathbf{o}$ aggregates multi-modal observations over the past $T_o$ steps
and $\mathbf{a} \in \mathbb{R}^{H \times d_a}$ is an expert action chunk of
horizon $H$ with per-step action dimensionality $d_a$. Flow matching learns a
velocity field $v_\theta$ that transports a \emph{source sample}
$\mathbf{z}_0 \in \mathbb{R}^{H \times d_a}$, drawn from a \emph{source
distribution} $q_0$, to the target action $\mathbf{a}$ along a linear
interpolation path:
\begin{equation}
    \mathbf{x}_t = (1 - t)\,\mathbf{z}_0 + t\,\mathbf{a}, \qquad
    \mathbf{u}_t = \mathbf{a} - \mathbf{z}_0,
    \label{eq:fm_setup}
\end{equation}
where $\mathbf{z}_0 \sim q_0$, $t \in [0, 1]$ is the flow time and $\mathbf{u}_t$ is the ground-truth
velocity along the path. The generator is trained to regress $\mathbf{u}_t$
under the observation conditioning $\mathbf{z}_{\text{cond}}$, and at inference
generates an action by drawing $\mathbf{z}_0 \sim q_0$ and integrating
$v_\theta$ from $t = 0$ to $t = 1$. Within this formulation, action generation is specified by three ingredients: the source distribution $q_0$, the velocity field $v_\theta$, and the integration scheme. Existing policies fix $q_0$ to an observation-independent $\mathcal{N}(\mathbf{0}, \mathbf{I})$; we instead treat $q_0$ as the design dimension of interest, leaving $v_\theta$ and the integration scheme unchanged.


\subsection{LeaP: A Proprioception-Conditioned Gaussian Prior}
\label{ssec:prior}
As illustrated in Fig.~\ref{fig:overview}, at each control step, the observation $\mathbf{o}$ is encoded into a visual feature $\mathbf{z}_{\text{vis}}$ and a proprioceptive feature $\mathbf{z}_s$.
LeaP replaces the standard source distribution
$\mathcal{N}(\mathbf{0}, \mathbf{I})$ with a proprioception-conditioned diagonal
Gaussian shared across the $H$ time steps of the action chunk:
\begin{equation}
    q_\phi(\mathbf{z}_0 \mid \mathbf{z}_s) = \prod_{h=1}^{H} \mathcal{N}\!\left(
        \mathbf{z}_{0,h};\, \boldsymbol{\mu}_\phi(\mathbf{z}_s),\,
        \mathrm{diag}(\boldsymbol{\sigma}^2_\phi(\mathbf{z}_s))
    \right),
    \qquad
    [\boldsymbol{\mu}_\phi,\, \log\boldsymbol{\sigma}^2_\phi]
    = g_\phi(\mathbf{z}_s) \in \mathbb{R}^{2 d_a}.
    \label{eq:prior}
\end{equation}
The prior network $g_\phi: \mathbb{R}^{d_s} \to \mathbb{R}^{2 d_a}$ is a
lightweight two-layer MLP whose output is split into a mean head and a
log-variance head, each of dimension $d_a$ (architecture details in
Appendix~\ref{app:arch_details}). The same
$(\boldsymbol{\mu}_\phi, \boldsymbol{\sigma}^2_\phi)$ parameterize the
per-step Gaussian for all $H$ time steps of the action chunk, reflecting the
prior's role of anchoring the chunk in a state-dependent neighborhood;
fine-grained per-step structure is left to the generator. We sample from this prior using the reparameterization
trick~\cite{kingma2014vae}, drawing an independent $\boldsymbol{\epsilon}_h$
for each time step:
\begin{equation}
    \mathbf{z}_{0,h} = \boldsymbol{\mu}_\phi(\mathbf{z}_s)
    + \boldsymbol{\sigma}_\phi(\mathbf{z}_s) \odot \boldsymbol{\epsilon}_h,
    \label{eq:reparam}
\end{equation}
where $\boldsymbol{\epsilon}_h \sim \mathcal{N}(\mathbf{0}, \mathbf{I}_{d_a})$ $(h = 1, \dots, H)$, so that both $\boldsymbol{\mu}_\phi$ and $\boldsymbol{\sigma}^2_\phi$ are
learned end-to-end from gradients flowing through $\mathbf{z}_0$. Conditioning
$g_\phi$ on $\mathbf{z}_s$ alone---rather than on the full feature
$\mathbf{z}_{\text{cond}} = [\mathbf{z}_{\text{vis}};\, \mathbf{z}_s]$
that $v_\theta$ receives---reflects the prior's role of anchoring
$\mathbf{z}_0$ in a state-dependent neighborhood of the action chunk;
finer-grained scene-level disambiguation within that neighborhood is left to
the generator. We examine the empirical consequences of this division of inputs, as well as alternative parameterizations of $q_\phi$, in Sec.~\ref{ssec:modality_ablation} and Sec.~\ref{ssec:form_ablation}. The sampled $\mathbf{z}_0$ is then passed to a flow-matching generator whose velocity field $v_\theta$, conditioned on the full feature $\mathbf{z}_{\text{cond}}$, integrates the linear path of Eq.~\eqref{eq:fm_setup} from $t{=}0$ to $t{=}1$ to transport $\mathbf{z}_0$ to action $\mathbf{a}$. LeaP imposes no structural constraint on $v_\theta$: the prior only specifies the distribution from which $\mathbf{z}_0$ is sampled, leaving the generator architecture and inference solver unchanged. We instantiate $v_\theta$ as a flow-matching network with adaptive layer normalization~\cite{dit} for both time and observation conditioning, and further verify in Sec.~\ref{ssec:backbone_ablation} that the same prior can be plugged into a diffusion-bridge generator with consistent gains.

\subsection{Joint Training Objective}
\label{ssec:losses}
The prior network $g_\phi$, projection heads $\phi_z, \phi_a$, and the
generator $v_\theta$ are optimized jointly under
\begin{equation}
    \mathcal{L} = \mathcal{L}_\text{flow}
    + \beta\, \mathcal{L}_\text{NLL}
    + \alpha\, \mathcal{L}_\text{align},
    \label{eq:total_loss}
\end{equation}
where $\mathcal{L}_\text{NLL}$ supervises the conditional density of
$q_\phi$, $\mathcal{L}_\text{align}$ supervises instance-level
$(\mathbf{z}_{0,i}, \mathbf{a}_i)$ correspondence in a learned embedding
space, and $\mathcal{L}_\text{flow}$ supervises $v_\theta$ while backpropagating to $g_\phi$ via Eq.~\eqref{eq:reparam}.
\paragraph{Density supervision} Treating each per-step action $\mathbf{a}_h$ in the expert chunk as a sample from the per-step Gaussian in Eq.~\eqref{eq:prior}, we supervise $g_\phi$ by
the negative log-likelihood, averaged over the $H$ time steps:
\begin{equation}
    \mathcal{L}_\text{NLL} = \mathbb{E}_{\mathcal{D}}\!\left[
        \frac{1}{2H} \sum_{h=1}^{H} \sum_{j=1}^{d_a} \left(
            \log \sigma^2_{\phi,j}(\mathbf{z}_s)
            + \frac{(a_{h,j} - \mu_{\phi,j}(\mathbf{z}_s))^2}{\sigma^2_{\phi,j}(\mathbf{z}_s)}
        \right)
    \right].
    \label{eq:nll}
\end{equation}

The precision-weighted squared error couples $\boldsymbol{\mu}_\phi$ and
$\boldsymbol{\sigma}^2_\phi$: residuals are penalized inversely by the
predicted variance, while $\log \sigma^2_{\phi,j}$ regularizes against
arbitrary inflation.

\paragraph{Instance-level alignment} We further supervise paired correspondence between source samples and
actions in a learned embedding space, using a symmetric contrastive loss in
the CLIP style~\cite{radford2021clip, oord2018cpc}:
\begin{equation}
    \mathcal{L}_\text{align} = \tfrac{1}{2}\!\left(
        \mathcal{L}_{z \to a} + \mathcal{L}_{a \to z}
    \right),
    \quad
    \mathcal{L}_{z \to a} = -\mathbb{E}_i \log
    \frac{\exp\!\left(\mathrm{sim}(\phi_z(\mathbf{z}_{0,i}), \phi_a(\mathbf{a}_i))/\tau\right)}
    {\sum_{k=1}^{B} \exp\!\left(\mathrm{sim}(\phi_z(\mathbf{z}_{0,i}), \phi_a(\mathbf{a}_k))/\tau\right)},
    \label{eq:align}
\end{equation}
with projections $\phi_z, \phi_a$, cosine similarity
$\mathrm{sim}(\mathbf{u}, \mathbf{v}) = \mathbf{u}^\top \mathbf{v} /
(\|\mathbf{u}\|\|\mathbf{v}\|)$, and temperature $\tau$. The reverse term
$\mathcal{L}_{a \to z}$ swaps the roles of $\mathbf{z}_0$ and $\mathbf{a}$.
This supervises a signal orthogonal to $\mathcal{L}_\text{NLL}$:
$\mathcal{L}_\text{NLL}$ matches the marginal $q_\phi(\mathbf{z}_0 \mid
\mathbf{z}_s)$ to the conditional action density, while
$\mathcal{L}_\text{align}$ couples each $\mathbf{z}_{0,i}$ to its specific
paired $\mathbf{a}_i$ at the sample level.

\paragraph{Flow matching loss} The generator regresses the target velocity along the interpolation path:
\begin{equation}
    \mathcal{L}_\text{flow} = \mathbb{E}_{(\mathbf{o}, \mathbf{a}, t,
    \boldsymbol{\epsilon})}\!\left[
        \big\| v_\theta(\mathbf{x}_t, \mathbf{z}_\text{cond}, t)
        - (\mathbf{a} - \mathbf{z}_0) \big\|^2
    \right].
    \label{eq:flow}
\end{equation}
Since $\mathbf{z}_0$ is reparameterized through $g_\phi$ (cf. Eq.~\eqref{eq:reparam}),
$\mathcal{L}_\text{flow}$ backpropagates to $\boldsymbol{\mu}_\phi$ and
$\boldsymbol{\sigma}^2_\phi$ alongside $v_\theta$, providing a third
supervision signal to $g_\phi$ on top of the two explicit objectives above.

\section{Experiments}
\label{sec:experiments}

\subsection{Experimental Setup}
\label{ssec:setup}


We evaluate LeaP on 15 RoboTwin~\citep{chen2025robotwin} bimanual manipulation tasks in simulation, spanning pick-and-place, articulated-object manipulation, tool use, and precise bimanual coordination, and on three real-world Franka Research 3 tasks spanning picking (\emph{Pick Cube}), articulated manipulation (\emph{Close Box}), and deformable-object handling (\emph{Pick \& Place Sandbags}). Simulation uses 50 expert demonstrations and 100 evaluation rollouts per task; real-world experiments use 100 teleoperated demonstrations and 20 evaluation rollouts per task with randomized object positions. In the main comparison, each method is trained and evaluated with 3 random seeds.

We compare LeaP against four baselines representing distinct choices of source distribution. \textbf{A2A}~\citep{jia2026action} and \textbf{NoPrior} isolate the effect of the source under an identical conditional SimpleFlowNet generator, using a state-derived deterministic latent and $\mathcal{N}(\mathbf{0},\mathbf{I})$ respectively; \textbf{VITA}~\citep{gao2025vita} uses the same network in its official conditioning-free form with a visual latent source; and \textbf{BridgePolicy}~\citep{bridgepolicy} jointly redesigns the source and generator dynamics through a diffusion bridge. For fair comparison, all methods share the same DP3 PointNet encoder~\citep{dp3,qi2017pointnet}: A2A and VITA follow their official implementations with this encoder as the only architectural change, while BridgePolicy, which has no public code, is reimplemented from the original paper (architecture and hyperparameters in Appendix~\ref{app:impl}). Streaming Flow Policy~\citep{jiang2025streaming} generates one action at a time rather than a full chunk, leaving no chunk-level source to compare, so we omit it as a baseline.


\subsection{Main Results}
\label{ssec:main_results}

Per-task success rates of the five methods on the 15 RoboTwin tasks are summarized in Table~\ref{tab:main}. LeaP attains the highest average success rate (81.6\%) and ranks first on the majority of individual tasks, outperforming A2A~\citep{jia2026action}, VITA~\citep{gao2025vita}, NoPrior, and BridgePolicy~\citep{bridgepolicy} by 6.5 to 25.5 percentage points. The largest gap, 25.5 pp over NoPrior, is particularly telling: NoPrior shares the same encoder, generator, and flow-matching pipeline as LeaP, and differs only in replacing the learned source prior with $\mathcal{N}(\mathbf{0},\mathbf{I})$. The 25.5 pp gap thus reflects the learnable prior as a whole rather than incidental architectural changes. A2A and VITA, which use deterministic observation-informed sources, fall between NoPrior and LeaP. Together, these comparisons suggest that the performance gain is not merely a consequence of moving away from the standard Gaussian source, but of doing so with a learnable, distributional prior.

\subsection{Efficiency Analysis}
\label{ssec:efficiency}

\begin{wrapfigure}{r}{0.54\linewidth}
  \centering
  \vspace{-1.5em}
  \includegraphics[width=\linewidth]{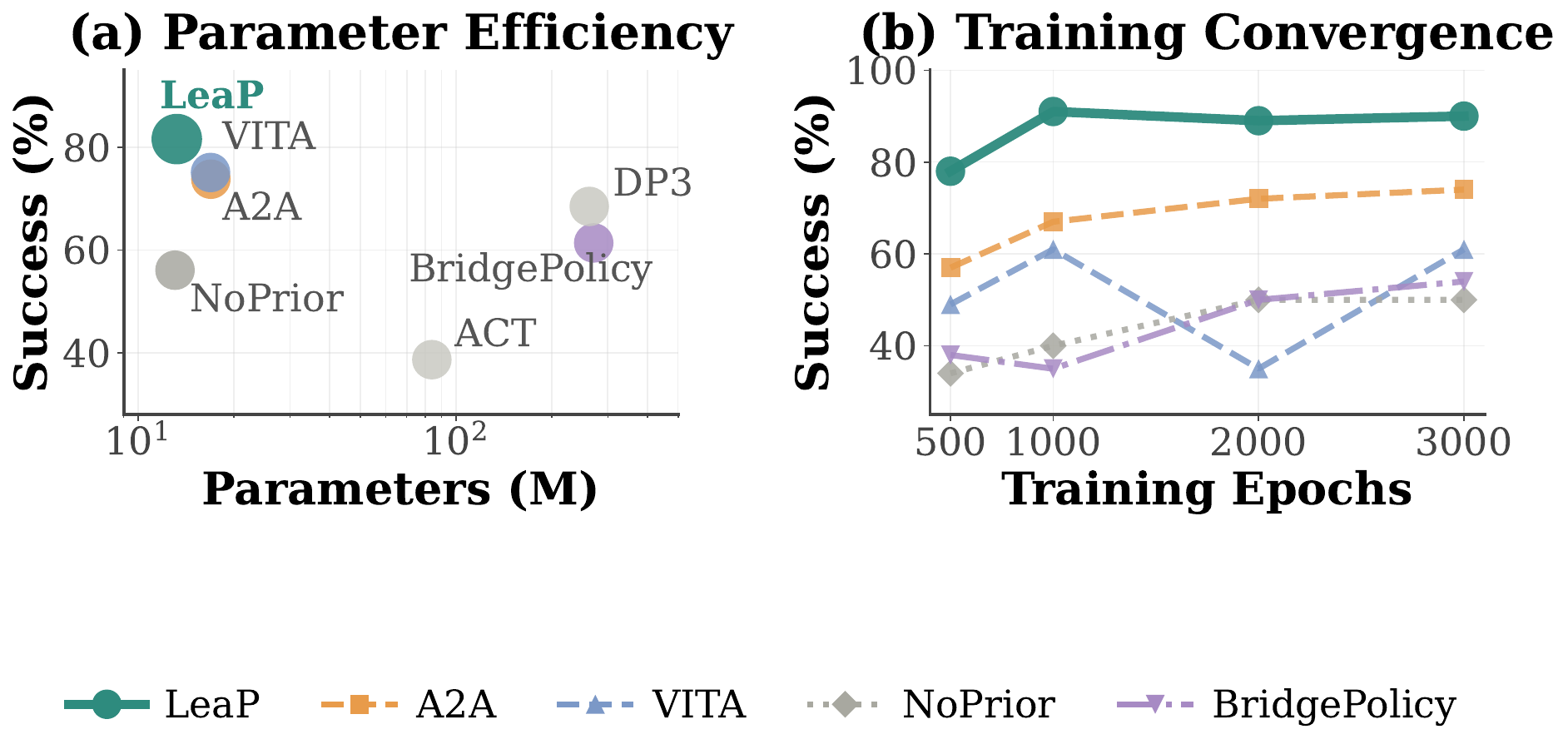}
  \caption{\textbf{Efficiency analysis.}
  (a) Parameter count versus 15-task average success rate (3-seed mean, matching Table~\ref{tab:main}).
  (b) Success rate on \texttt{open\_laptop} across training epochs (seed~0).}
  \label{fig:efficiency}
  \vspace{-1em}
\end{wrapfigure}



We analyze efficiency along two axes: parameter count and training convergence. Fig.~\ref{fig:efficiency}(a) plots the 15-task average success rate against model size. LeaP uses 13.22M parameters, of which the prior head $g_\phi$ accounts for only 0.21M (about 1.6\%), a negligible overhead over the architecture-matched NoPrior. A2A and VITA add latent-action modules and use 28\% more parameters than LeaP, yet score lower; ACT, DP3, and BridgePolicy rely on larger Transformer or UNet backbones (6--20$\times$ more parameters) but still do not match LeaP. Convergence shows the same advantage: on \texttt{open\_laptop} (Fig.~\ref{fig:efficiency}(b), seed~0), LeaP reaches 91\% within 1000 epochs---surpassing the final success of A2A, VITA, NoPrior, and BridgePolicy at 3000 epochs---and stays stable near 90\% thereafter, while A2A converges more slowly and VITA and BridgePolicy fluctuate more (e.g., VITA drops from 61\% to 35\% between epochs 1000 and 2000). Together, these results indicate that the source distribution is a low-overhead lever that improves policy performance while training with fewer parameters and in fewer epochs.

\newcommand{\msb}[2]{\textbf{#1}{\scriptsize$\pm$#2}} 
\newcommand{\msu}[2]{\underline{#1}{\scriptsize$\pm$#2}} 
\newcommand{\ms}[2]{#1{\scriptsize$\pm$#2}}            

\begin{table}[t]
\centering
\caption{Per-task success rates (\%) on 15 RoboTwin manipulation tasks, reported as mean$\pm$std over 3 seeds (100 rollout episodes per seed). \textbf{Bold} marks the best result per task, \underline{underline} the second best. LeaP achieves the highest average success rate, surpassing all baselines by a substantial margin.}
\label{tab:main}
\footnotesize
\setlength{\tabcolsep}{3pt}
\begin{tabular}{lccccc}
\toprule
Task & \textbf{LeaP} & A2A & VITA & \makecell{NoPrior\\($\mathcal{N}(\mathbf{0},\mathbf{I})$)} & BridgePolicy \\
\midrule
beat\_block\_hammer     & \msb{94.0}{2.6}  & \ms{89.3}{2.3}  & \msu{89.7}{2.9}  & \ms{46.7}{4.0}  & \ms{79.3}{3.1}  \\
dump\_bin\_bigbin       & \msb{92.3}{6.7}  & \ms{88.0}{6.0}  & \msu{89.0}{1.7}  & \ms{70.7}{6.4}  & \ms{77.3}{9.5}  \\
click\_bell             & \msb{100.0}{0.0} & \ms{94.7}{1.5}  & \msu{98.0}{2.6}  & \ms{92.0}{13.9} & \ms{77.7}{4.0}  \\
handover\_block         & \msu{91.7}{1.5}  & \msb{92.3}{2.1} & \ms{83.0}{6.6}   & \ms{38.7}{11.6} & \ms{60.3}{3.8}  \\
place\_phone\_stand     & \msb{75.0}{0.0}  & \ms{61.0}{4.6}  & \msu{64.3}{2.1}  & \ms{50.3}{4.0}  & \ms{37.0}{1.0}  \\
lift\_pot               & \msb{98.3}{1.5}  & \ms{92.7}{4.2}  & \msu{97.0}{0.0}  & \ms{87.0}{2.0}  & \ms{91.3}{2.5}  \\
move\_playingcard\_away & \msu{75.0}{4.4}  & \ms{72.0}{2.0}  & \msb{76.3}{4.0}  & \ms{59.0}{5.0}  & \ms{67.3}{5.0}  \\
open\_laptop            & \msb{86.0}{6.9}  & \msu{70.3}{4.0} & \ms{62.7}{3.8}   & \ms{46.3}{3.5}  & \ms{50.3}{3.5}  \\
pick\_diverse\_bottles  & \msb{77.0}{6.1}  & \ms{43.3}{5.5}  & \msu{63.0}{4.6}  & \ms{37.3}{8.1}  & \ms{34.0}{3.0}  \\
click\_alarmclock       & \msb{97.0}{1.0}  & \msu{89.7}{4.0} & \ms{88.3}{8.0}   & \ms{81.3}{6.1}  & \ms{67.3}{10.0} \\
place\_empty\_cup       & \ms{86.7}{4.2}   & \msb{88.3}{3.5} & \msu{88.0}{1.7}  & \ms{74.7}{21.7} & \ms{67.0}{3.6}  \\
press\_stapler          & \msb{71.3}{9.5}  & \ms{56.0}{5.3}  & \ms{56.7}{5.8}   & \msu{68.0}{8.7} & \ms{49.7}{6.7}  \\
place\_can\_basket      & \msb{81.3}{2.1}  & \ms{73.3}{7.6}  & \ms{73.3}{11.7}  & \ms{46.3}{17.8} & \msu{77.0}{2.6} \\
stack\_blocks\_two      & \msu{33.7}{0.6}  & \ms{31.0}{2.0}  & \msb{35.0}{7.5}  & \ms{16.7}{3.8}  & \ms{22.0}{3.6}  \\
place\_object\_basket   & \msu{64.3}{4.0}  & \msb{65.7}{3.2} & \ms{62.3}{5.5}   & \ms{27.0}{6.6}  & \ms{63.3}{5.5}  \\
\midrule
\textbf{Average}        & \msb{81.6}{0.4}  & \ms{73.8}{2.2}  & \msu{75.1}{0.9}  & \ms{56.1}{2.5}  & \ms{61.4}{1.3}  \\
\bottomrule
\end{tabular}
\end{table}

\begin{figure}[t]
\centering
\includegraphics[width=0.95\linewidth]{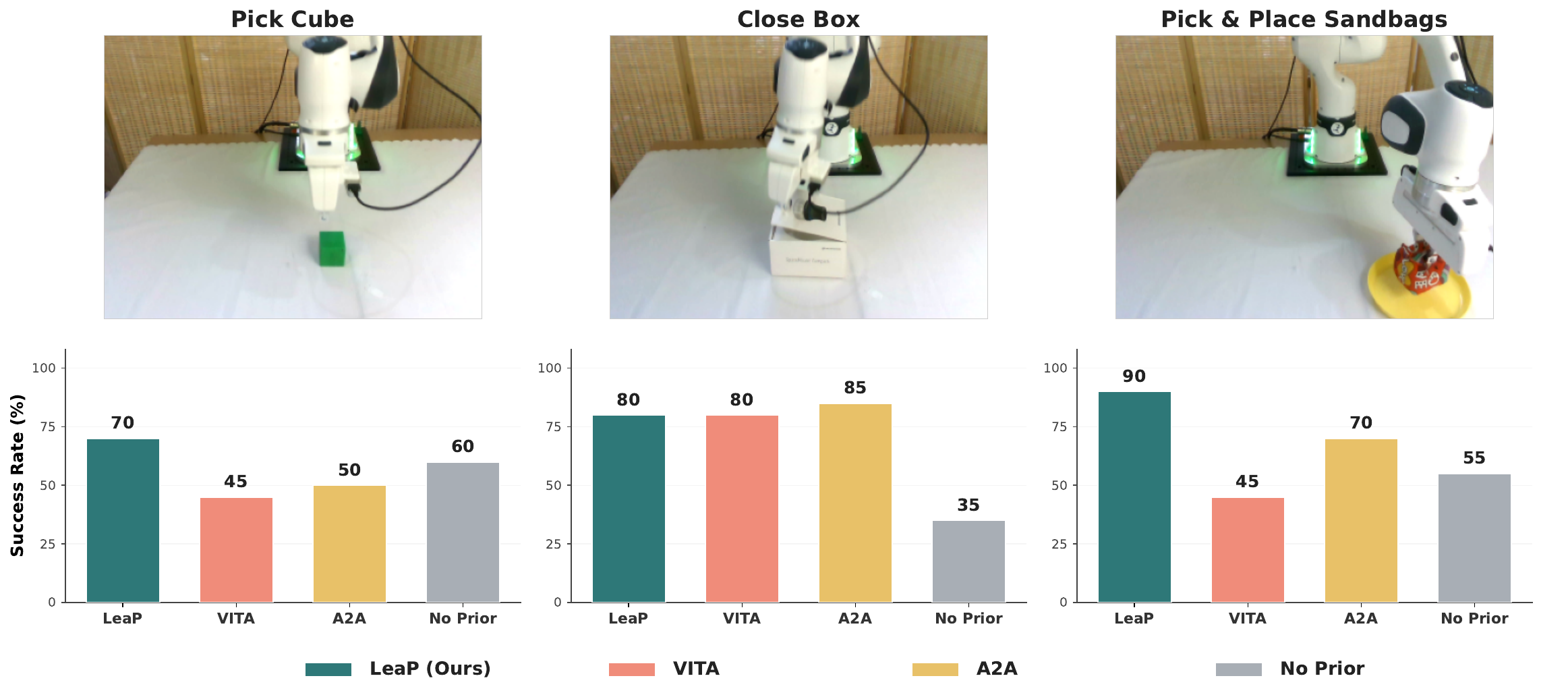}
\vspace{-0.5em}
\captionsetup{skip=4pt}
\caption{Real-robot evaluation on a Franka Research 3. 
\textit{Top:} task settings for the three manipulation tasks. 
\textit{Bottom:} per-task success rates of each method. LeaP 
achieves the highest average success rate (80.0\%), outperforming A2A (68.3\%), VITA (56.7\%), and NoPrior (46.7\%).}
\label{fig:real_robot}
\end{figure}

\subsection{Real-Robot Experiments}
\label{ssec:real_robot}

Fig.~\ref{fig:real_robot} reports the real-robot results. LeaP achieves the highest average success rate, 80.0\%, outperforming A2A (68.3\%), VITA (56.7\%), and NoPrior (46.7\%) by 11.7, 23.3, and 33.3 percentage points, respectively. The main trend from simulation carries over: LeaP remains the top performer and the architecture-matched NoPrior remains last in both settings, reinforcing the simulation finding that the gain is attributable to the 
learned prior itself (Sec.~\ref{ssec:main_results}).

\section{Ablation Studies}
\label{sec:ablation}

We ablate three core design dimensions of LeaP: \textbf{(i)} what observation modality the prior should be conditioned on (Sec.~\ref{ssec:modality_ablation}), \textbf{(ii)} how the prior should be parameterized as a probability distribution (Sec.~\ref{ssec:form_ablation}), and \textbf{(iii)} how the prior should be supervised during training (Sec.~\ref{ssec:supervision_ablation}). We further verify that LeaP plugs into different generators (Sec.~\ref{ssec:backbone_ablation}). Average success rates are summarized in Fig.~\ref{fig:ablation}; per-task results and additional source-distribution visualizations are provided in Appendices~\ref{app:ablation_tables} and~\ref{app:source_vis}, respectively. All variants are evaluated on three RoboTwin tasks (\texttt{pick\_diverse\_bottles}, \texttt{open\_laptop}, \texttt{handover\_block}) with 100 rollout episodes per task; details follow Appendix~\ref{app:impl}. Each variant uses seed~0, so we read the ablations as trends rather than precise estimates.

\subsection{What Should Condition the Prior?}
\label{ssec:modality_ablation}

\begin{wrapfigure}{r}{0.51\linewidth}
  \centering
  \vspace{-2.5em}
  \includegraphics[width=\linewidth]{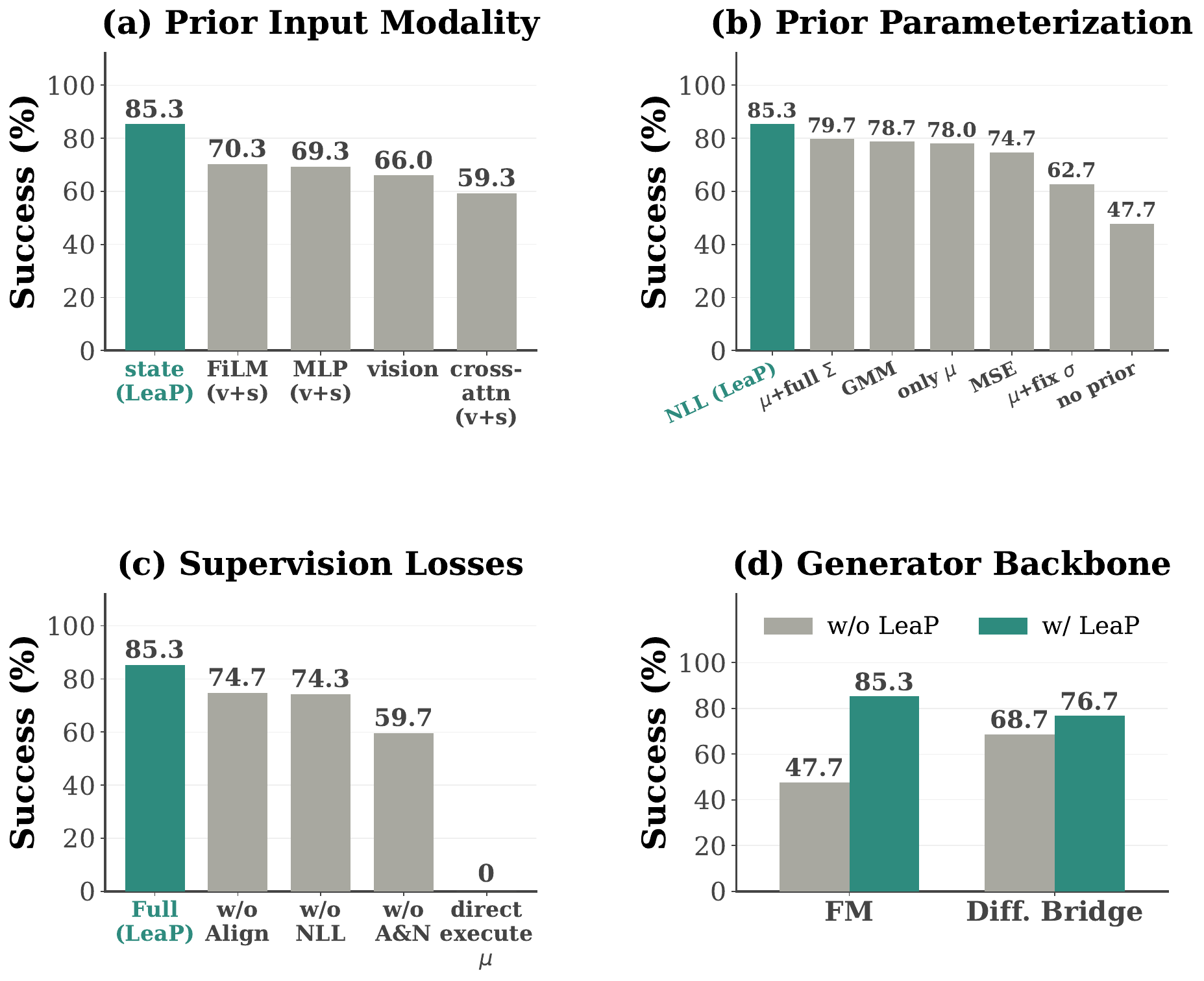}
  \caption{3-task average success rate across four design axes of LeaP---prior input (a), parameterization (b), supervision (c), and generator backbone (d). \emph{v+s} denotes vision+state fusion in (a).}
  \label{fig:ablation}
  \vspace{-1.5em}
\end{wrapfigure}

We vary the input to $g_\phi$ across proprioceptive state, vision, and three vision+state fusions (FiLM, MLP, cross-attention), with $\mathbf{z}_\text{cond}$ held fixed. Fig.~\ref{fig:ablation}(a) shows that state alone achieves the best average success rate (85.3\%), outperforming all other variants by 15--26 percentage points. The drop is uniform across the three fusion strategies, suggesting that the degradation is associated with routing visual features into the prior, rather than with a particular fusion choice. Per-task results are provided in Appendix~\ref{app:modality_table}. Fig.~\ref{fig:source_modality} shows that the state-only prior places source samples near the target action, whereas vision-bearing variants shift the source distribution away. This supports conditioning the prior on proprioception, while leaving visual information to the generator.

\begin{figure}[t]
\centering
\includegraphics[width=\linewidth]{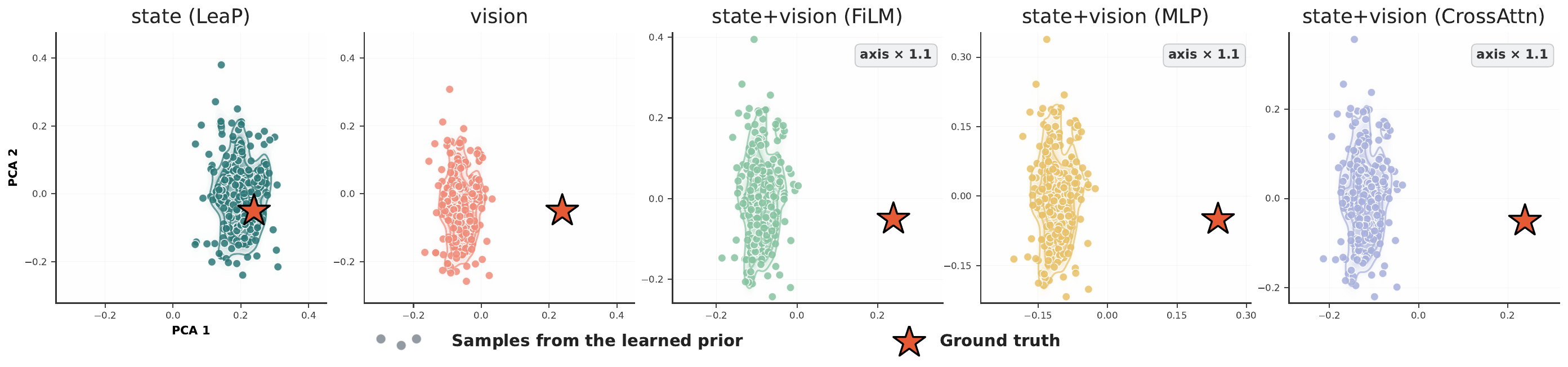}
\vspace{-1.5em}
\captionsetup{skip=6pt}
\caption{\textbf{Source distributions across prior-input modalities on \texttt{handover\_block}.} 
For the same target action $\mathbf{a}$ (red star), we draw $K{=}200$ source samples $\mathbf{z}_0$ from each prior variant and project them to 2D with PCA. 
The state-only (LeaP) prior places samples near the target, whereas variants that ingest visual features shift the source distribution away.}
\label{fig:source_modality}
\vspace{-0.6em}
\end{figure}


\subsection{What Form Should the Prior Take?}
\label{ssec:form_ablation}

We compare seven prior parameterizations: the NLL-trained diagonal Gaussian used by LeaP, $\mu$+full $\boldsymbol{\Sigma}$, only $\mu$, a Gaussian mixture with $k{=}8$ components, an MSE-supervised point estimate, $\mu$+fixed $\sigma$ ($\sigma{=}1$), and the no-prior standard-Gaussian baseline. Fig.~\ref{fig:ablation}(b) reports the 3-task averages. LeaP achieves the highest success rate (85.3\%), and all learnable priors outperform the no-prior baseline. However, more expressive parameterizations do not help here: both $\mu$+full $\boldsymbol{\Sigma}$ and GMM underperform the diagonal Gaussian (LeaP). Variants without a state-adaptive variance also trail LeaP: only $\mu$ reaches 78.0\%, MSE reaches 74.7\%, and $\mu$+fixed $\sigma$ drops to 62.7\%. Notably, $\mu$+fixed $\sigma$ falls 15.3 points below only $\mu$, suggesting that the gain does not come from stochasticity alone, but from learning state-adaptive uncertainty. Per-task results are in Appendix~\ref{app:form_table}. Fig.~\ref{fig:source_form} shows the corresponding source distributions: LeaP places samples near the target action with a tight spread, whereas the standard Gaussian source lies far from the target, and fixed-variance or higher-capacity variants either spread too broadly or shift mass away. These results point to state-adaptive uncertainty that anchors generation locally, rather than expressivity or stochasticity alone.

\begin{figure}[t]
\centering
\includegraphics[width=\linewidth]{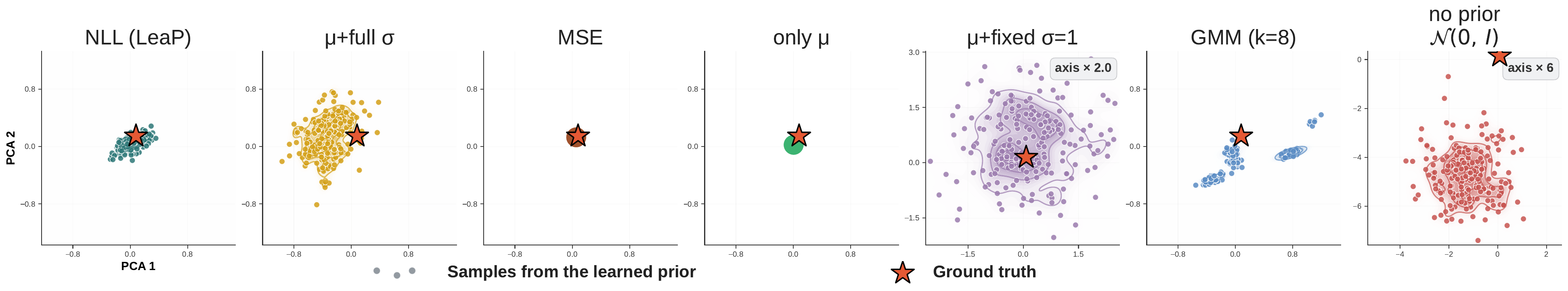}
\captionsetup{skip=4pt}
\caption{\textbf{Source distributions across prior parameterizations on \texttt{handover\_block}.} 
For the same target action $\mathbf{a}$ (red star), we draw $K{=}200$ source samples $\mathbf{z}_0$ from each stochastic prior variant and project them to 2D with PCA; deterministic variants (\textit{MSE} and \textit{only $\mu$}) are shown as single-point estimates. 
NLL (LeaP) places the source distribution near the target with a tight spread, while other stochastic forms become overly diffuse or shift probability mass away.}
\label{fig:source_form}
\vspace{-0.6em}
\end{figure}

\subsection{What Should Supervise the Prior?}
\label{ssec:supervision_ablation}

The three terms in $\mathcal{L}$ play distinct roles. We isolate their
contributions by removing $\mathcal{L}_\text{align}$,
$\mathcal{L}_\text{NLL}$, or both, while keeping the prior input and
parameterization fixed. Fig.~\ref{fig:ablation}(c) reports the 3-task averages.
Removing $\mathcal{L}_\text{align}$ or $\mathcal{L}_\text{NLL}$ drops success
from 85.3\% to 74.7\% and 74.3\%, respectively, suggesting that the two losses
provide complementary supervision. Removing both leaves $g_\phi$ trained only
through the reparameterized flow loss, reaching 59.7\%---above the no-prior
baseline (47.7\%) but well below the full model. Directly executing
$\mu_\phi(\mathbf{z}_s)$ as the action yields 0\%, confirming that the prior
provides a source for the generator rather than replacing it. Per-task results
are in Appendix~\ref{app:supervision_table}. Additional source-distribution
visualizations in Appendix~\ref{app:source_vis} show the same trend: the full
objective forms compact source distributions near the target action, whereas
removing explicit prior supervision weakens source--target alignment.

\subsection{Generality across Generators}
\label{ssec:backbone_ablation}


Finally, we test whether the same LeaP prior remains effective across
generator families. We compare the flow-matching backbone with a
diffusion-bridge generator~\citep{unidb,unidbpp}. The bridge formulation requires
a non-trivial source: initializing from $\mathcal{N}(\mathbf{0},\mathbf{I})$
yields $0\%$ success. We therefore use LeaP's deterministic mean
$\boldsymbol{\mu}_\phi(\mathbf{z}_s)$ as the minimal feasible source,
paired with the same bridge backbone as BridgePolicy. Fig.~\ref{fig:ablation}(d)
and Table~\ref{tab:backbone_app} show that LeaP improves both generators:
from $47.7\%$ to $85.3\%$ on flow matching ($+37.6$ pp), and from $68.7\%$
to $76.7\%$ on diffusion bridge ($+8.0$ pp), with the latter isolating the
contribution of state-adaptive variance. LeaP thus acts as a reusable source-distribution module rather
than a mechanism tied to a specific generator.

\section{Conclusion}
\label{sec:conclusion}

We presented LeaP, a learnable source-prior module for generative robot policies. LeaP models the starting point of action generation as a learnable distribution---a proprioception-conditioned diagonal Gaussian over the action chunk---providing a calibrated, stochastic source before generative refinement, while leaving the downstream generator architecture and inference solver unchanged. Across simulation and real-world manipulation tasks, LeaP improves over prior methods and transfers across both flow-matching and diffusion-bridge generators, indicating that source-distribution design complements generator dynamics rather than replacing them. These results suggest that \emph{where action generation begins} is a reusable design axis for generative robot policies: a lightweight, learnable source distribution can serve as an effective lever for improving policy performance without redesigning the generator.

\section{Limitations}
\label{sec:limitations}

We parameterize the source as a diagonal Gaussian, which does not model correlations across action dimensions; our preliminary attempt with a full covariance matrix did not yield gains, and structured parameterizations that capture correlations at lower cost---such as low-rank-plus-diagonal covariance or normalizing-flow priors---remain to be explored. Our evaluation is also restricted to tabletop manipulation; whether the gains persist on vision-language-action models and more complex robot embodiments such as dexterous hands remains untested.


\clearpage


\bibliography{example}  

\clearpage
\appendix
\section*{Appendix}

This appendix is organized in four parts. Appendix~\ref{app:impl} gives the full
implementation of LeaP and all baselines on RoboTwin, from the shared training
protocol down to method-specific differences. Appendix~\ref{app:task_gallery} shows
the initial scene of each of the 15 RoboTwin simulation tasks.
Appendix~\ref{app:ablation_tables} reports per-task numbers for every ablation in
Sec.~\ref{sec:ablation}, and Appendix~\ref{app:source_vis} provides the matching
source-distribution visualizations, following the same set of design axes.

\section{Implementation Details}
\label{app:impl}

All methods are trained and evaluated under a common protocol and share the same
DP3 PointNet point-cloud encoder~\citep{dp3,qi2017pointnet}. We first describe the
shared setup and hyperparameters (Appendix~\ref{app:shared_hparams}), then the
network architectures (Appendix~\ref{app:arch_details}), the few settings that
differ across methods (Appendix~\ref{app:method_settings}), the diffusion-bridge
baseline, which requires separate treatment (Appendix~\ref{app:bridge_settings}),
and finally model-size and backbone comparisons (Appendix~\ref{app:params}).

\subsection{Shared Setup and Hyperparameters}
\label{app:shared_hparams}

For all RoboTwin simulation experiments, each task uses 50 expert demonstrations,
and every method is trained for 3000 epochs and evaluated over 100 rollout
episodes per task. The ablation studies in Sec.~\ref{sec:ablation} use three
tasks---\texttt{pick\_diverse\_bottles}, \texttt{open\_laptop}, and
\texttt{handover\_block}---under the same protocol. Table~\ref{tab:shared_hparams}
lists the hyperparameters shared across all methods.

\begin{table}[!htbp]
\centering
\caption{Shared hyperparameters for RoboTwin simulation experiments.}
\label{tab:shared_hparams}
\small
\setlength{\tabcolsep}{8pt}
\begin{tabular}{lc}
\toprule
Hyperparameter & Value \\
\midrule
Batch size & 256 \\
Learning rate & $1\times10^{-4}$ \\
LR scheduler & cosine \\
LR warmup steps & 500 \\
Observation steps $n_{\mathrm{obs}}$ & 3 \\
Action execution steps $n_{\mathrm{act}}$ & 6 \\
Prediction horizon & 8 \\
Training epochs & 3000 \\
Optimizer & AdamW \\
EMA & \checkmark \\
\bottomrule
\end{tabular}
\end{table}

\subsection{Architecture Details}
\label{app:arch_details}

We detail the architecture of our method (LeaP) and its architecture-matched
counterpart (NoPrior); the baselines are described at the end. Both LeaP and
NoPrior are built on the shared DP3 PointNet encoder above and an MLP-based
SimpleFlowNet velocity field, in which the timestep embedding and the observation
feature drive adaLN-style (FiLM) modulation---per-layer scale, shift, and gate
applied to a LayerNorm'd hidden state. The two are identical except that NoPrior
removes the prior head and samples the source from $\mathcal{N}(\mathbf{0},
\mathbf{I})$. The A2A baseline shares this same SimpleFlowNet generator, so that
LeaP, NoPrior, and A2A isolate the effect of the source distribution under an
identical flow-matching backbone; VITA uses the same network in its official
conditioning-free configuration.

\paragraph{LeaP prior network $g_\phi$.}
The prior head is a two-layer MLP applied to the proprioceptive feature
$\mathbf{z}_s$: an input LayerNorm, a Linear layer expanding to a $512$-d hidden
representation with ReLU activation and dropout ($p{=}0.1$), and a final Linear
layer producing $2 d_a$ outputs that are split into the mean
$\boldsymbol{\mu}_\phi \in \mathbb{R}^{d_a}$ and log-variance
$\log \boldsymbol{\sigma}^2_\phi \in \mathbb{R}^{d_a}$. The same
$(\boldsymbol{\mu}_\phi, \boldsymbol{\sigma}^2_\phi)$ are shared across the $H$
time steps of the action chunk.

\paragraph{LeaP projection heads $\phi_z, \phi_a$.}
The two projection heads used in $\mathcal{L}_\text{align}$ are two-layer MLPs
(Linear--ReLU--Linear) mapping to a shared $128$-d embedding space, applied
independently to the flattened source sample $\mathbf{z}_0$ and the flattened
action chunk $\mathbf{a}$ (each of dimension $H d_a$).

\paragraph{Baselines.} A2A~\citep{jia2026action} and VITA~\citep{gao2025vita} are
based on their official implementations, with the visual encoder replaced by the
shared DP3 PointNet encoder as the only architectural change; we refer to the
original papers for their architectural and loss details. BridgePolicy~\citep{bridgepolicy}
has no public code and is reimplemented from the original paper (Appendix~E of
\citet{bridgepolicy}), including its multi-modal fusion module and semantic
aligner, using a ConditionalUNet1D backbone with downsampling dimensions
$[512,1024,2048]$ and kernel size $5$.

\subsection{Method-specific Settings}
\label{app:method_settings}

Given the shared protocol and architectures above, the methods differ only in the
source distribution, the auxiliary losses, and---for BridgePolicy---the backbone.
Table~\ref{tab:method_settings} summarizes these differences; LeaP and NoPrior are
identical except that NoPrior removes the learned prior head and samples from the
standard Gaussian source. For the flow-matching schedule, LeaP, A2A, and NoPrior
use a conditional CFM path with linear interpolation ($\sigma_{\mathrm{FM}}{=}0.0$),
while VITA uses its official exact-OT-CFM path ($\sigma_{\mathrm{FM}}{=}0.0$); all
four use forward-Euler integration with 6 function evaluations at inference.

\begin{table}[!htbp]
\centering
\caption{Method-specific settings for RoboTwin simulation experiments. Batch size is 256 for all methods (Table~\ref{tab:shared_hparams}). A2A and VITA additionally use their native action-autoencoder losses at the official default weights (consistency $1.0$, reconstruction $0.5$); BridgePolicy uses a bridge loss ($1.0$).}
\label{tab:method_settings}
\small
\begin{tabular*}{\linewidth}{@{\extracolsep{\fill}}lllll@{}}
\toprule
Method & NFE & Source type & Main network & Prior / align loss \\
\midrule
LeaP & 6 & proprioception Gaussian & SimpleFlowNet & NLL $1.0$, Align $0.3$ \\
NoPrior & 6 & $\mathcal{N}(\mathbf{0},\mathbf{I})$ & SimpleFlowNet & -- \\
VITA & 6 & visual latent & SimpleFlowNet (cond.-free) & -- \\
A2A & 6 & state latent & SimpleFlowNet & -- \\
BridgePolicy & 10 & vision+state bridge & ConditionalUNet1D & Align $0.3$ \\
\bottomrule
\end{tabular*}
\end{table}

\subsection{BridgePolicy Reimplementation}
\label{app:bridge_settings}

BridgePolicy differs from the other baselines in both its source and its
generator dynamics, and no official code is available; we therefore reimplement it
following the architecture and hyperparameters of the original paper (Appendix~E
of \citet{bridgepolicy}), including the multi-modal fusion module and semantic
aligner. All shared settings (50 demonstrations per task, 3000 training epochs,
100 evaluation rollouts) match the protocol used for the other baselines for fair
comparison. BridgePolicy-specific hyperparameters are listed in
Table~\ref{tab:bridge_hparams}.

\begin{table}[!htbp]
\centering
\caption{BridgePolicy-specific hyperparameters.}
\label{tab:bridge_hparams}
\small
\setlength{\tabcolsep}{8pt}
\begin{tabular}{lc}
\toprule
Hyperparameter & Value \\
\midrule
Training discretization \texttt{num\_train\_timesteps} & 100 \\
Inference NFE & 10 \\
\texttt{bridge\_lambda\_square} & 0.05 \\
\texttt{bridge\_gamma} & $1\times10^{6}$ \\
\texttt{bridge\_schedule} & cosine \\
\texttt{bridge\_eps} & 0.01 \\
Solver & UniDB++~\citep{unidbpp} mean-ODE \\
\bottomrule
\end{tabular}
\end{table}

\subsection{Model Size and Backbones}
\label{app:params}

Table~\ref{tab:params_app} reports the parameter count and backbone of LeaP and
all baselines, supporting the efficiency analysis in Sec.~\ref{ssec:efficiency}.
LeaP uses fewer parameters than every external baseline while attaining the
highest success rate (Table~\ref{tab:main}); the learnable prior head adds only
$0.21$M over its architecture-matched ablation NoPrior ($13.01$M vs.\ $13.22$M).

\begin{table}[!htbp]
\centering
\caption{Parameter counts and backbones of LeaP and the baselines. LeaP uses
fewer parameters than every external baseline; the prior head adds only $0.21$M over the
architecture-matched ablation NoPrior.}
\label{tab:params_app}
\small
\setlength{\tabcolsep}{12pt}
\begin{tabular}{lrl}
\toprule
Method & Params & Backbone \\
\midrule
NoPrior              & 13.01M  & MLP \\
\textbf{LeaP (Ours)} & 13.22M  & MLP \\
A2A                  & 16.92M  & MLP \\
VITA                 & 16.88M  & MLP \\
ACT                  & 83.89M  & Transformer \\
DP3                  & 262.43M & UNet \\
BridgePolicy         & 270.99M & UNet \\
\bottomrule
\end{tabular}
\end{table}

\section{Simulation Task Gallery}
\label{app:task_gallery}

\begin{figure}[t]
    \centering
    \includegraphics[width=\linewidth]{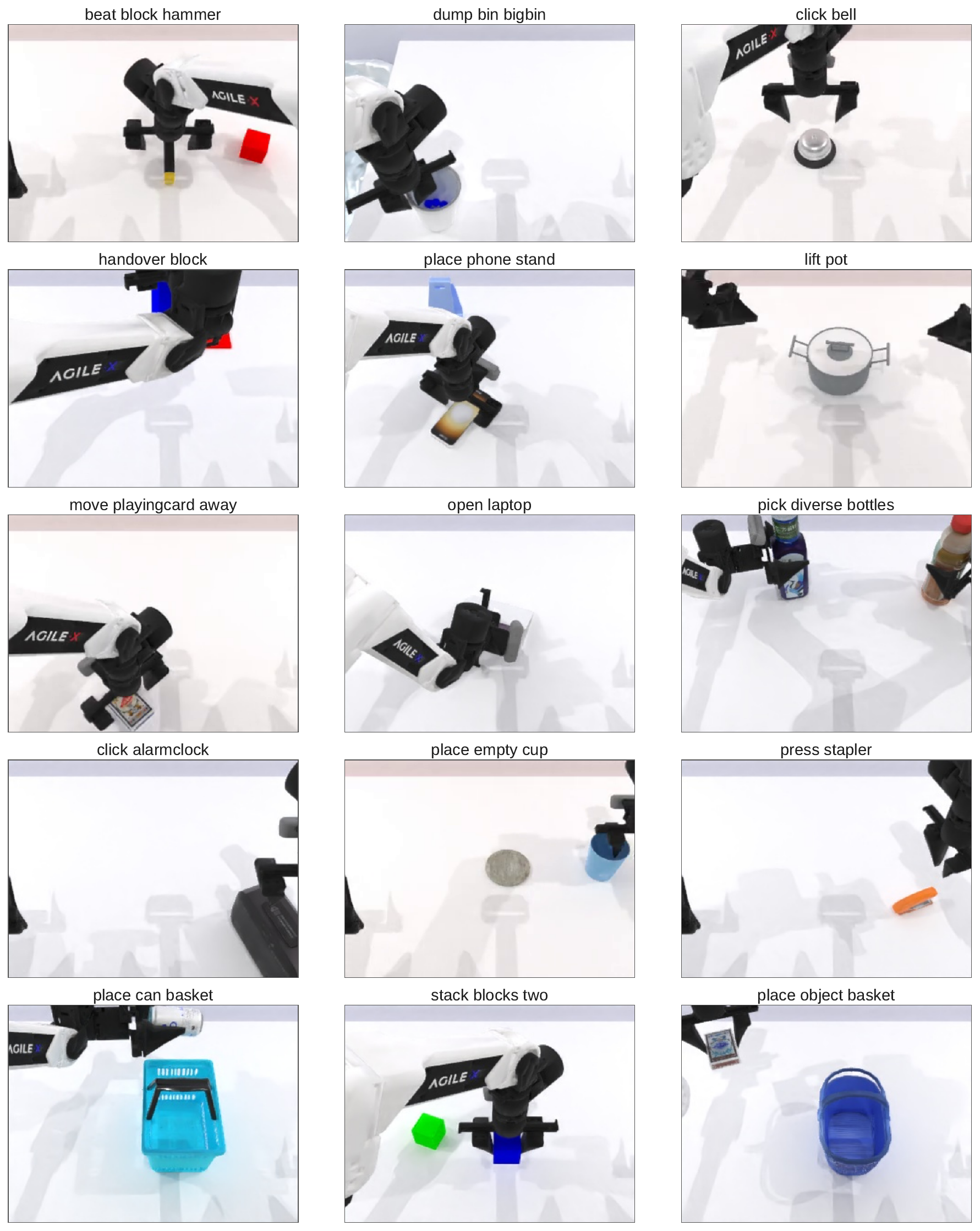}
    \caption{\textbf{The 15 RoboTwin simulation tasks.} Each panel shows a
representative initial configuration and is labeled with the task name. Object
positions are randomized across rollout episodes.}
    \label{fig:task_gallery}
\end{figure}

Figure~\ref{fig:task_gallery} shows the initial scene of each of the 15 RoboTwin 2.0
manipulation tasks used in our simulation experiments
(Sec.~\ref{ssec:main_results}). The tasks span pick-and-place, articulated-object
manipulation, tool use, and bimanual coordination, covering varied contact and
precision requirements.

\section{Per-task Ablation Results}
\label{app:ablation_tables}

This section reports the per-task numbers behind the averaged results in
Fig.~\ref{fig:ablation}, following the same four design axes as
Sec.~\ref{sec:ablation}: prior-input modality
(Appendix~\ref{app:modality_table}), prior parameterization
(Appendix~\ref{app:form_table}), prior supervision
(Appendix~\ref{app:supervision_table}), and generator generality
(Appendix~\ref{app:backbone_table}). Unless otherwise noted, all variants are
evaluated on \texttt{pick\_diverse\_bottles}, \texttt{open\_laptop}, and
\texttt{handover\_block} with 100 rollout episodes per task.

\subsection{Prior-input Modality}
\label{app:modality_table}

The prior head input is varied while the generator's full conditioning
$\mathbf{z}_{\mathrm{cond}}$ is held fixed (Table~\ref{tab:modality_app}).

\begin{table}[!htbp]
\centering
\caption{Per-task success rates (\%) for the prior-input modality ablation.}
\label{tab:modality_app}
\small
\setlength{\tabcolsep}{6pt}
\begin{tabular}{lcccc}
\toprule
Prior Input & pick\_diverse\_bottles & open\_laptop & handover\_block & \textbf{Avg} \\
\midrule
\textbf{state (LeaP)} & 74 & \textbf{90} & \textbf{92} & \textbf{85.3} \\
vision & 66 & 77 & 55 & 66.0 \\
vision+state (FiLM) & \textbf{77} & 82 & 52 & 70.3 \\
vision+state (MLP) & 72 & 80 & 56 & 69.3 \\
vision+state (CrossAttn) & 76 & 69 & 33 & 59.3 \\
\bottomrule
\end{tabular}
\end{table}

\subsection{Prior Parameterization}
\label{app:form_table}

The prior input is held fixed at proprioceptive state while its parameterization
is varied (Table~\ref{tab:form_app}).

\begin{table}[!htbp]
\centering
\caption{Per-task success rates (\%) for the prior-parameterization ablation.}
\label{tab:form_app}
\small
\setlength{\tabcolsep}{6pt}
\begin{tabular}{lcccc}
\toprule
Prior Form & pick\_diverse\_bottles & open\_laptop & handover\_block & \textbf{Avg} \\
\midrule
\textbf{NLL (LeaP)} & \textbf{74} & \textbf{90} & 92 & \textbf{85.3} \\
$\mu$+full $\boldsymbol{\Sigma}$ & 68 & 85 & 86 & 79.7 \\
GMM ($k{=}8$) & 70 & 85 & 81 & 78.7 \\
only $\mu$ & 63 & 78 & \textbf{93} & 78.0 \\
MSE & 50 & 87 & 87 & 74.7 \\
$\mu$+fixed $\sigma$ & 47 & 63 & 78 & 62.7 \\
no prior & 41 & 50 & 52 & 47.7 \\
\bottomrule
\end{tabular}
\end{table}

\subsection{Prior Supervision}
\label{app:supervision_table}

The input modality and prior form are held fixed at proprioceptive state and the
diagonal Gaussian prior, respectively, while supervision terms are removed
(Table~\ref{tab:supervision_app}).

\begin{table}[!htbp]
\centering
\caption{Per-task success rates (\%) for the prior-supervision ablation.}
\label{tab:supervision_app}
\small
\setlength{\tabcolsep}{6pt}
\begin{tabular}{lcccc}
\toprule
Supervision & pick\_diverse\_bottles & open\_laptop & handover\_block & \textbf{Avg} \\
\midrule
\textbf{Full (LeaP)} & 74 & \textbf{90} & \textbf{92} & \textbf{85.3} \\
w/o Align & 68 & 86 & 70 & 74.7 \\
w/o NLL & 63 & 79 & 81 & 74.3 \\
w/o Align \& NLL & \textbf{79} & 79 & 21 & 59.7 \\
direct $\boldsymbol{\mu}_\phi$ execution & 0 & 0 & 0 & 0.0 \\
\bottomrule
\end{tabular}
\end{table}

\subsection{Generator Generality}
\label{app:backbone_table}

We attach the same LeaP prior to two generator families
(Table~\ref{tab:backbone_app}). The diffusion-bridge variants use LeaP's
deterministic mean $\boldsymbol{\mu}_\phi$ as source (``Bridge'') or the full
proprioception-conditioned Gaussian (``Bridge+LeaP''), both paired with the same
diffusion-bridge formulation and backbone as BridgePolicy.

\begin{table}[!htbp]
\centering
\caption{Per-task success rates (\%) for the generator-generality ablation. FM: flow matching; Bridge: diffusion bridge.}
\label{tab:backbone_app}
\small
\setlength{\tabcolsep}{6pt}
\begin{tabular}{lcccc}
\toprule
Task & FM & FM+LeaP & Bridge ($\boldsymbol{\mu}_\phi$) & Bridge+LeaP \\
\midrule
pick\_diverse\_bottles & 41 & \textbf{74} & 64 & \textbf{74} \\
open\_laptop           & 50 & \textbf{90} & 57 & 62 \\
handover\_block        & 52 & 92 & 85 & \textbf{94} \\
\midrule
\textbf{Average}       & 47.7 & \textbf{85.3} & 68.7 & 76.7 \\
\bottomrule
\end{tabular}
\end{table}

\section{Additional Source-distribution Visualizations}
\label{app:source_vis}

To assess the consistency of the qualitative trends in Sec.~\ref{sec:ablation},
we visualize source distributions over 10 randomly sampled target actions on
\texttt{handover\_block}, organized along the same three prior-design axes as
Appendix~\ref{app:ablation_tables}: prior-input modality
(Appendix~\ref{app:source_vis_modality}), prior parameterization
(Appendix~\ref{app:source_vis_form}), and prior supervision
(Appendix~\ref{app:source_vis_supervision}). For each axis we draw source samples
from each prior variant. The red star denotes the target action. Since some variants produce distributions far from the target, the plotting axes are rescaled independently when necessary for visibility; the axis multiplier is shown in each panel.

\subsection{Prior-input Modality}
\label{app:source_vis_modality}

\begin{figure}[p]
    \centering
    \includegraphics[height=0.92\textheight,width=\linewidth,keepaspectratio]{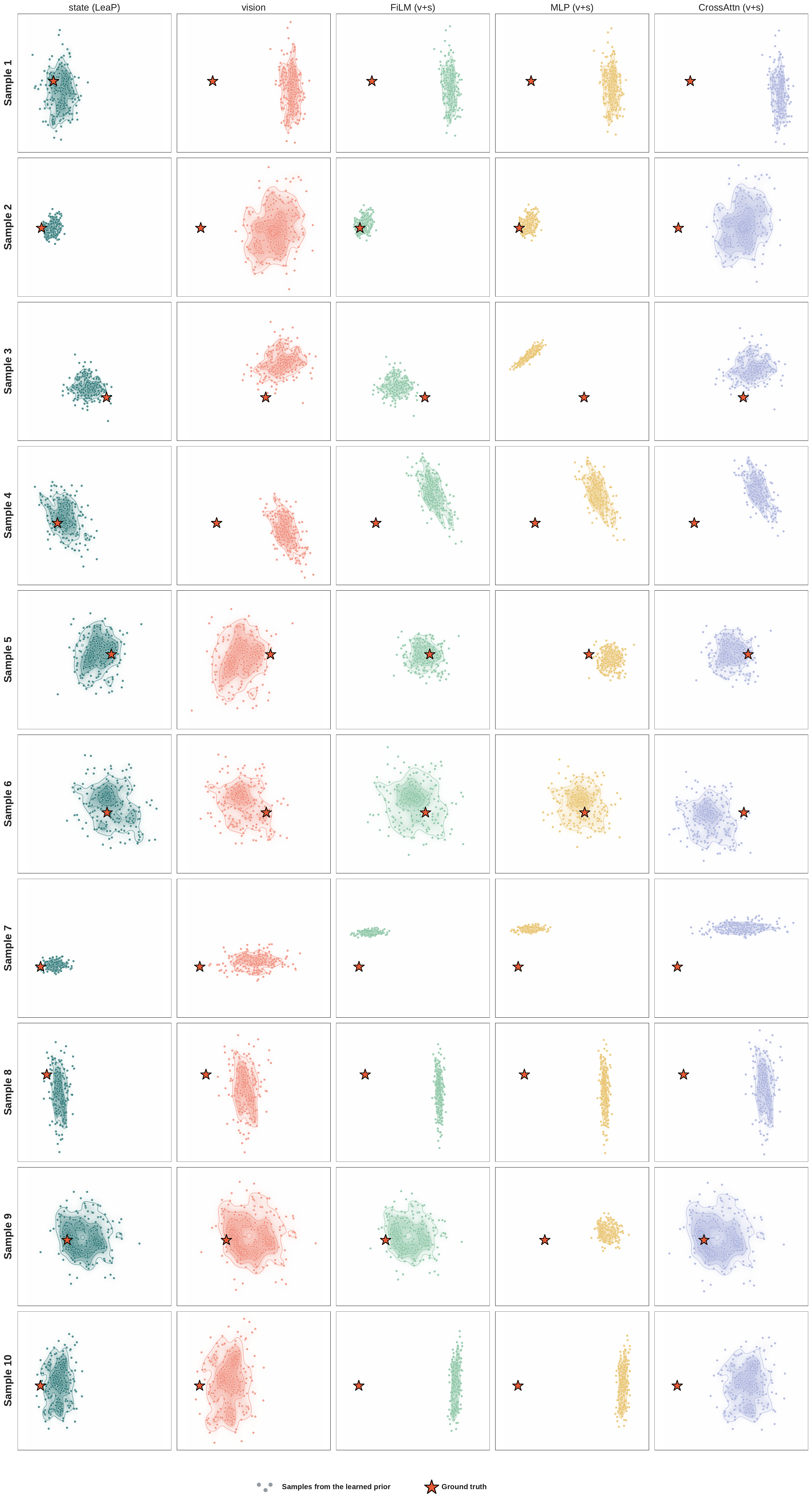}
    \caption{\textbf{Additional source-distribution visualizations for the prior-input modality ablation.}
    We visualize 10 randomly selected target actions on \texttt{handover\_block}. The red star denotes the target action. The state-only prior used by LeaP  places source samples closer to the target than vision-only or vision--state fusion variants.}
    \label{fig:source_modality_app}
\end{figure}

Across the 10 targets (Fig.~\ref{fig:source_modality_app}), the state-conditioned
prior used by LeaP places source samples close to the target action, whereas
vision-only and vision--state fusion variants more often shift the source cloud
away. This matches the success-rate trend in Sec.~\ref{ssec:modality_ablation}:
proprioception is the better signal for anchoring the source, while visual
information is better left to the downstream generator.

\subsection{Prior Parameterization}
\label{app:source_vis_form}

\begin{figure}[p]
    \centering
    \includegraphics[height=0.92\textheight,width=\linewidth,keepaspectratio]{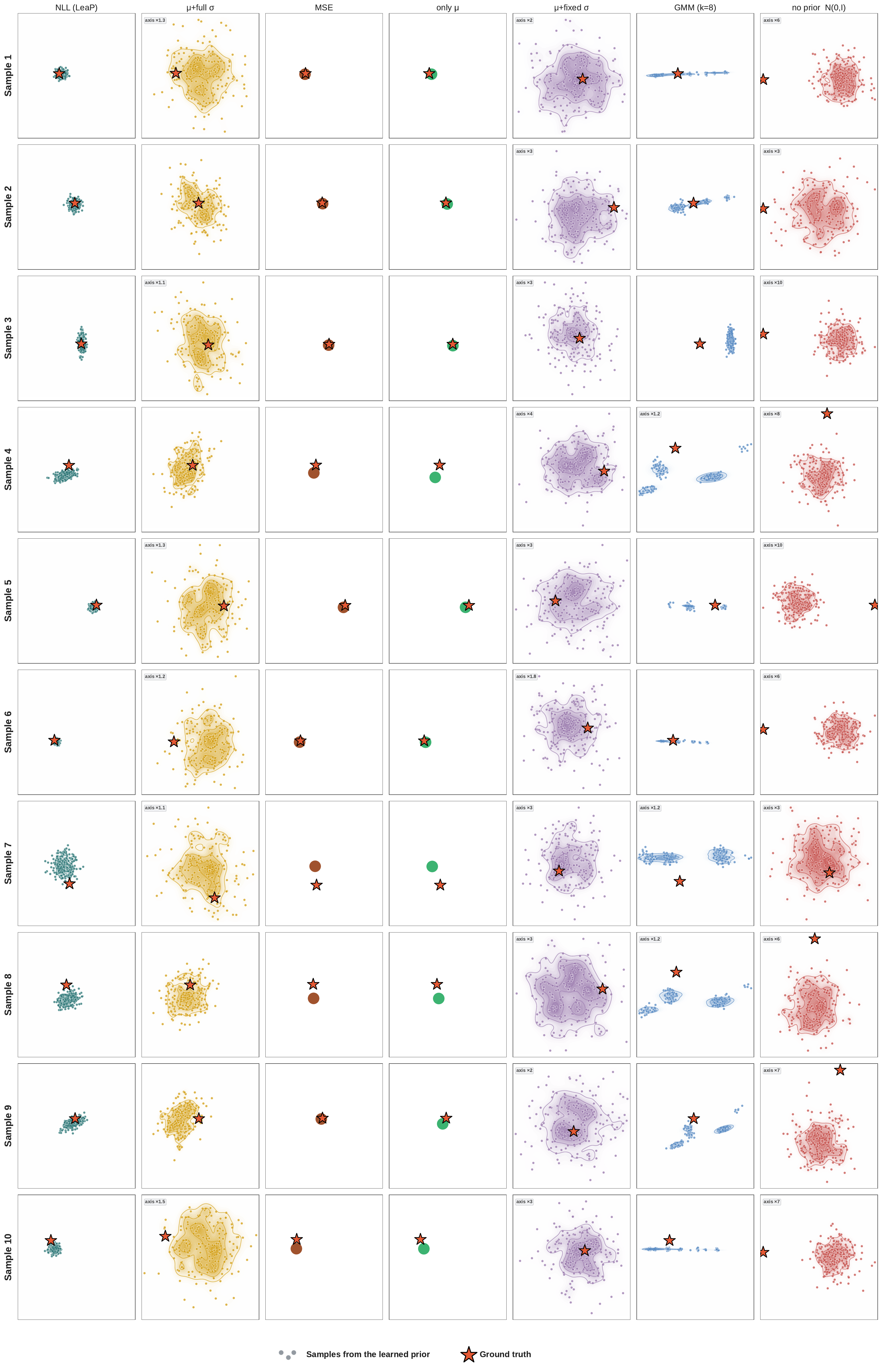}
    \caption{\textbf{Additional source-distribution visualizations for the prior-parameterization ablation.}
    We visualize 10 randomly selected target actions on \texttt{handover\_block}. The NLL-trained diagonal Gaussian used by LeaP places source samples near the target with a compact spread, whereas the standard Gaussian, fixed-variance, and higher-capacity alternatives are more likely to be diffuse or misaligned.}
    \label{fig:source_form_app}
\end{figure}

Across the 10 targets (Fig.~\ref{fig:source_form_app}), the NLL-trained diagonal
Gaussian used by LeaP produces compact source distributions near the
target, whereas the standard Gaussian source lies far from the target and
fixed-variance or higher-capacity alternatives become diffuse or shift mass
away. This agrees with Table~\ref{tab:form_app}: state-adaptive uncertainty
matters more than simply increasing prior expressivity or injecting stochasticity.

\subsection{Prior Supervision}
\label{app:source_vis_supervision}

\begin{figure}[p]
    \centering
    \includegraphics[height=0.92\textheight,width=\linewidth,keepaspectratio]{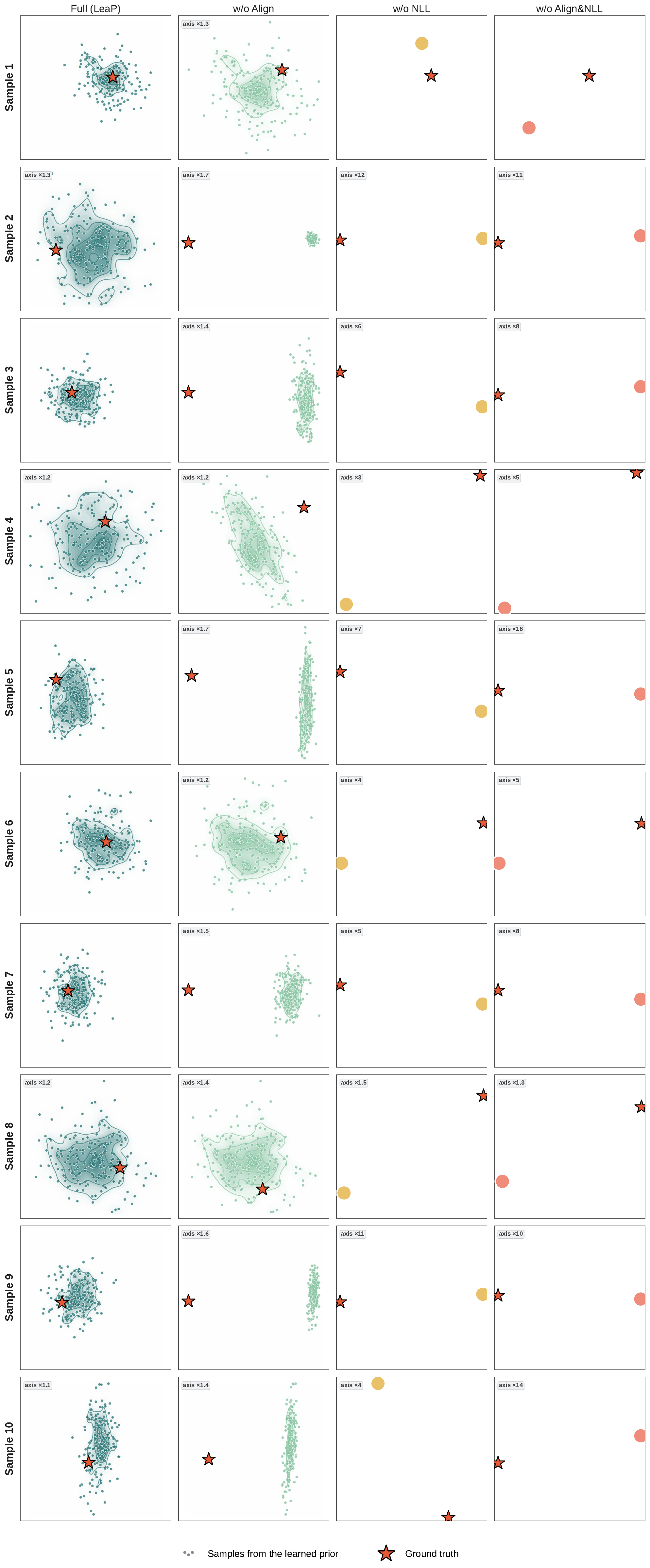}
    \caption{\textbf{Additional source-distribution visualizations for the prior-supervision ablation.}
    We visualize 10 randomly selected target actions on \texttt{handover\_block}. The full LeaP objective produces source clouds close to the target, whereas removing explicit prior-supervision terms degrades the alignment between the source distribution and the target action.}
    \label{fig:source_supervision_app}
\end{figure}

Across the 10 targets (Fig.~\ref{fig:source_supervision_app}), the full objective
yields source clouds near the target; removing $\mathcal{L}_\text{align}$ weakens
concentration, while removing $\mathcal{L}_\text{NLL}$ or both explicit
prior-supervision terms moves the source far from the target. This explains, at
the distribution level, the quantitative drops in Table~\ref{tab:supervision_app}:
explicit density and alignment supervision are complementary for learning a
calibrated source distribution.

\end{document}